\documentclass[letterpaper, 10 pt, conference]{ieeeconf}
\IEEEoverridecommandlockouts
\usepackage{times}
\usepackage{xcolor}

\usepackage{times}
\usepackage{helvet}
\usepackage{courier}
\usepackage{amsfonts}
\usepackage{amsmath}
\usepackage{amssymb}
\usepackage{graphicx}
\usepackage{graphics}
\usepackage{epsfig}
\usepackage{mathptmx}
\usepackage{url}
\usepackage[l]{floatflt}
\usepackage{rotating}
\usepackage{subfigure}
\usepackage{multicol}
\usepackage{algorithm}
\usepackage{algorithmic}
\usepackage{verbatim}
\usepackage{listings}
\usepackage{chngpage}
\usepackage{wrapfig,lipsum}
\usepackage{booktabs}
\usepackage{threeparttable}
\usepackage{multirow}
\DeclareMathAlphabet{\mathcal}{OMS}{cmsy}{m}{n}

\floatname{algorithm}{Procedure}

\usepackage{color,url}
\usepackage{hyperref}

\long\def\c#1{{\footnotesize{\fontfamily{pcr}\selectfont{#1}}}}

\title{\LARGE \bf Robot Representation and Reasoning with Knowledge \\
from Reinforcement Learning}

\author{Keting Lu$^1$, Shiqi Zhang$^2$, Peter Stone$^3$, Xiaoping Chen$^1$\\
$^1$ University of Science and Technology of China,
$^2$ SUNY Binghamton,
$^3$ UT Austin\\
{\tt\small ktlu@mail.ustc.edu.cn; szhang@cs.binghamton.edu; pstone@cs.utexas.edu; xpchen@ustc.edu.cn}
}

\begin{document}

\maketitle
\thispagestyle{empty}
\pagestyle{empty}

\begin{abstract}
Reinforcement learning (RL) agents aim at learning by interacting with an environment, and are not designed for representing or reasoning with declarative knowledge.
Knowledge representation and reasoning (KRR) paradigms are strong in declarative KRR tasks, but are ill-equipped to learn from such experiences.
In this work, we integrate logical-probabilistic KRR with model-based RL, enabling agents to simultaneously reason with declarative knowledge and learn from interaction experiences.
The knowledge from humans and RL is unified and used for dynamically computing task-specific planning models under potentially new environments.
Experiments were conducted using a mobile robot working on dialog, navigation, and delivery tasks.
Results show significant improvements, in comparison to existing model-based RL methods.
\end{abstract}

%%%%%%%%%%%%%%%%%%%%%%%%%%%%%%%%%%%%%%%%%%%%%%%%%%%%%%%%%%
%%%%%%%%%%%%%%%%%%%%%%%%%%%%%%%%%%%%%%%%%%%%%%%%%%%%%%%%%%
%%%%%%%%%%%%%%%%%%%%%%%%%%%%%%%%%%%%%%%%%%%%%%%%%%%%%%%%%%

\section{Introduction}

Knowledge representation and reasoning (KRR) and reinforcement learning (RL) are two important research areas in artificial intelligence (AI) and have been applied to a variety of problems in robotics.
On the one hand, KRR research aims to concisely represent knowledge, and robustly draw conclusions with the knowledge (or generate new knowledge).
Knowledge in KRR is typically provided by human experts in the form of declarative rules.
Although KRR paradigms are strong in representing and reasoning with knowledge in a variety of forms, they are not designed for (and hence not good at) learning from experiences of accomplishing the tasks.
On the other hand, RL algorithms enable agents to learn by interacting with an environment, and RL agents are good at learning action policies from trial-and-error experiences toward maximizing long-term rewards under uncertainty, but they are ill-equipped to utilize declarative knowledge from human experts.
Motivated by the complementary features of KRR and RL, we aim at a framework that integrates both paradigms to enable agents (robots in our case) to simultaneously reason with declarative knowledge and learn by interacting with an environment.

Most KRR paradigms support the representation and reasoning of knowledge in logical form, e.g., the Prolog-style.
More recently, researchers have developed KRR paradigms that support both logical and probabilistic knowledge.
Examples include Markov Logic Network (MLN)~\cite{richardson2006markov}, P-log~\cite{baral2009probabilistic}, and Probabilistic Soft Logic (PSL)~\cite{bach2017hinge}.
Such logical-probabilistic KRR paradigms can be used for a variety of reasoning tasks.
We use P-log in this work to represent and reason with both human knowledge and the knowledge from RL.
When a task becomes available, our robot reasons at runtime to produce probabilistic transition systems for action policy generation.

Reinforcement learning (RL) algorithms can be used to help robots learn action policies from the experience of interacting with the real world~\cite{sutton1998reinforcement}.
There are at least two types of RL algorithms, namely model-based RL and model-free RL, depending on whether a world model is computed or not.
Model-free RL aims at directly computing the ``value'' of a state (or state-action pair), whereas the output of model-based RL includes a world model, and one can use planning algorithms to compute action policies with the world model.
We use model-based RL in this work, because the learned world model can be used to update the robot's declarative knowledge base and combined with human knowledge.

In this paper, we develop a framework (KRR-RL) that integrates logical-probabilistic KRR and model-based RL.
Transition probabilities that result from actions are learned via model-based RL, and then incorporated into the probabilistic reasoning module, which in turn (together with human knowledge) enables dynamic construction of efficient run-time task-specific planning models.
Our KRR-RL framework, for the first time, enables a robot to:
i) represent the probabilistic knowledge (i.e., world dynamics) learned from RL in declarative form;
ii) unify and reason with both human knowledge and the knowledge from RL; and
iii) compute policies at runtime by dynamically constructing task-oriented partial world models.
The above advantages are backed with experiments both in simulation and using real mobile robots conducting navigation, dialog, and delivery tasks.

\section{Related Work}
\label{sec:related}

This work is on integrating logical-probabilistic knowledge representation and reasoning (KRR) and model-based reinforcement learning (RL).
Related research areas include integrated logical KRR and RL, relational RL, and integrated KRR and probabilistic planning.

For example, logical KRR has previously been integrated with RL.
Action knowledge~\cite{mcdermott1998pddl,jiang2018empirical} has been used to reason about action sequences and help an RL agent explore only the states that can potentially contribute to achieving the ultimate goal~\cite{leonetti2016synthesis}.
As a result, their agents learn faster by avoiding choosing ``unreasonable'' actions.
A similar idea has been applied to domains with non-stationary dynamics~\cite{ferreira2017answer}.
More recently, task planning was used to interact with the high level of a hierarchical RL framework~\cite{yang2018peorl}.
The goal shared by these works is to enable RL agents to use knowledge to improve the performance in learning (e.g., to learn faster and/or avoid risky exploration).
However, the KRR capabilities of these methods are limited to \emph{logical} action knowledge.
By contrast, we use a logical-probabilistic KRR paradigm that can directly reason with probabilities learned from RL.

Relational RL (RRL) combines RL with relational reasoning~\cite{dvzeroski2001relational}.
Action models have been incorporated into RRL, resulting in a relational temporal difference learning method~\cite{asgharbeygi2006relational}.
Recently, RRL has been deployed for learning affordance relations that forbid the execution of specific actions~\cite{sridharan2017can}.
These RRL methods, including deep RRL~\cite{zambaldi2018relational}, exploit structural representations over states and actions in (only) current tasks.
In this research, our framework supports the KRR of factors beyond the ones in state and action representations, e.g., \emph{time} in navigation tasks (Section~\ref{sec:ins}).

The research area of integrated KRR and probabilistic planning is also related to this research.
Logical-probabilistic reasoning has been used to compute informative priors~\cite{zhang2015corpp} and world dynamics~\cite{zhang2017dynamically} for probabilistic planning.
An action language was used to compute a deterministic sequence of actions for robots, where individual actions are then implemented using probabilistic controllers~\cite{sridharan2015refinement}.
Recently, human-provided information has been incorporated into belief state representations to guide robot action selection~\cite{chitnis2018integrating}.
World models must be provided to these methods, where learning was not involved.

Finally, there are a number of robot reasoning and learning architectures~\cite{tenorth2013knowrob,oh2015toward,hanheide2017robot,khandelwal2017bwibots}, which are relatively complex, and support a variety of functionalities.
In comparison, we aim at a concise representation for robot KRR and RL capabilities.
To the best of our knowledge, this is the first work on a tightly coupled integration of logical-probabilistic KRR with model-based RL, where the KRR component supports the representation of and reasoning with the knowledge both from a human and from RL.

\section{Integrated KRR-RL Framework}
\label{sec:alg}

\begin{figure}[tb]
  \begin{center}
    \includegraphics[width=0.45\textwidth]{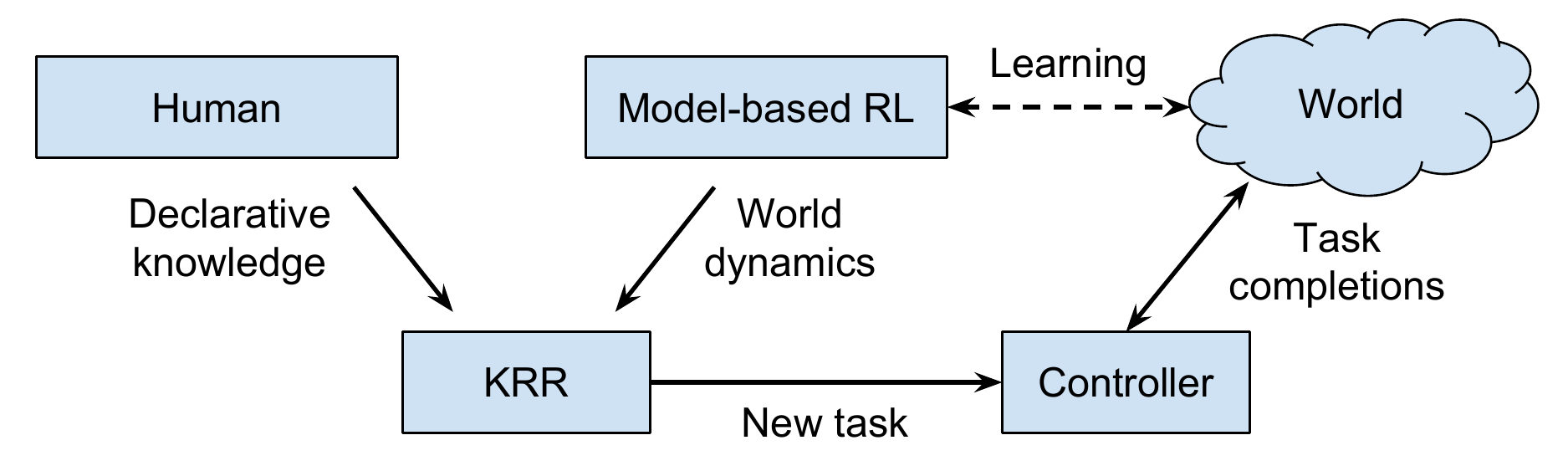}
  \end{center}
  \vspace{-1.5em}
  \caption{A pictorial overview of our KRR-RL framework. }
  \label{fig:overview}
  \vspace{-1.5em}
\end{figure}

Our unified framework for logical-probabilistic knowledge representation and reasoning (KRR) and model-based reinforcement learning (RL) is illustrated in Figure~\ref{fig:overview}.
The KRR includes both human knowledge and the knowledge learned from model-based RL.
When the robot is free, the robot arbitrarily selects goals (different navigation goals in our case) to work on, and learns the world dynamics, e.g., success rates and costs of navigation actions.
When a task becomes available, the KRR component dynamically constructs a partial world model (excluding unrelated factors), on which a task-oriented controller is computed using planning algorithms.
Human knowledge is on environment variables and their dependencies, e.g., navigation actions' success rates depend on current time and weather (laser sensors can be blinded in areas near east-facing windows in sunny mornings), while the robot must learn specific probabilities by interacting with the environment.

% \vspace{-1em}
\noindent
\textbf{\emph{Why integrated KRR-RL is needed? }}	
Consider an indoor robot navigation domain, where a robot wants to maximize the success rate of moving to goal positions through navigation actions.
\emph{Shall we include factors, such as time, weather, positions of human walkers, etc, into the state space?}
On the one hand, to ensure model completeness, the answer should be ``yes''.
Human walkers and sunlight reduce the success rates of the robot's navigation actions,
% e.g., strong sunlight (in sunny mornings near east-facing windows) probabilistically blinds the robot's laser sensor; human walkers probabilistically block the way;
and both can cause the robot irrecoverably lost.
On the other hand, to ensure computational feasibility, the answer is ``no''.
Modeling whether one specific grid cell being occupied by humans or not introduces one extra dimension in the state space, and doubles the state space size.
If we consider (only) ten such grid cells, the state space becomes $2^{10}\approx 1000$  times bigger.
As a result, RL practitioners frequently have to make a trade-off between model completeness and computational feasibility.
In this work, we aim at a framework that retains both model scalability and computational feasibility, i.e., the agent is able to learn within relatively small spaces while computing action policies accounting for a large number of domain variables.

\subsection{A General Procedure}

In factored spaces, state variables $\mathcal{V}=\{V_0,V_1,...,V_{n-1}\}$ can be split into two categories, namely endogenous variables $\mathcal{V}^{en}$ and exogenous variables $\mathcal{V}^{ex}$~\cite{chermack2004improving}, where $\mathcal{V}^{en}=\{V_0^{en},V_1^{en},...,V_{p-1}^{en}\}$ and $\mathcal{V}^{ex}=\{V_0^{ex},V_1^{ex},...,V_{q-1}^{ex}\}$.
In our integrated KRR-RL context, $\mathcal{V}^{en}$ is goal-oriented and includes the variables whose values the robot wants to actively change so as to achieve the goal; and $\mathcal{V}^{ex}$ corresponds to the variables whose values affect the robot's action outcomes, but the robot cannot (or does not want to) change their values.
Therefore, $\mathcal{V}^{en}$ and $\mathcal{V}^{ex}$ are both functions of task $\tau$.
Continuing the navigation example, robot position is an endogenous variable, and current time is an exogenous variable.
For each task, $\mathcal{V}=\mathcal{V}^{en} \cup \mathcal{V}^{ex}$ and $n=p+q$, and RL agents learn in spaces specified by $\mathcal{V}^{en}$.

The KRR component models $V$, their dependencies from human knowledge, and conditional probabilities on how actions change their values, as learned through model-based RL.
When a task arrives, the KRR component uses probabilistic rules to generate a task-oriented Markov decision process (MDP)~\cite{puterman1994markov}, which only contains a subset of $\mathcal{V}$ that are relevant to the current task, i.e., $\mathcal{V}^{en}$, and their transition probabilities.
Given this task-oriented MDP, a corresponding action policy is computed using value iteration or policy iteration.

\begin{algorithm}[tb]\small
  \caption{Learning in KRR-RL Framework}
  \label{alg:framework}
  \begin{algorithmic}[1]
    \REQUIRE{Logical rules $\Pi^L$;
        probabilistic rules $\Pi^P$;
        random variables $\mathcal{V}=\{V_0,V_1,...,V_{n-1}\}$;
        task selector $\Delta$; and
        guidance functions (from human knowledge) of $f^V(\mathcal{V}, \tau)$ and $f^A(\tau)$}
    \WHILE{Robot has no task}
        \STATE{$\tau \leftarrow \Delta()$: a task is heuristically selected}
        \STATE{$\mathcal{V}^{en} \leftarrow f^V(\mathcal{V}, \tau)$, and $\mathcal{V}^{ex} \leftarrow \mathcal{V} \setminus \mathcal{V}^{en}$}
    \STATE{$A \leftarrow f^A(\tau)$}
%     \STATE{Reason transition probability $\mathcal{P}$ of task $\tau^s$}
    \STATE{$\mathcal{M} \leftarrow \textit{Procedure-}\ref{alg:generatemodel}(\Pi^L, \Pi^P, \mathcal{V}^{en}, \mathcal{V}^{ex}, A)$}
    \STATE{Initialize agent: $agent\leftarrow$ \emph{R-Max}$(\mathcal{M})$}
    \STATE{RL \emph{agent} repeatedly works on task $\tau$, and keeps maintaining task model $\mathcal{M}'$, until policy convergence}
    \ENDWHILE
    \STATE{Use $\mathcal{M}'$ to update $\Pi^P$}
  \end{algorithmic}
\end{algorithm}

Our KRR-RL agent learns by interacting with an environment when there is no task assigned (Procedure~\ref{alg:framework}).
As soon as a task arrives, it uses the probabilities that are marked ``known'' to update the knowledge base, and start to focus on completing the task at hand (Procedure~\ref{alg:generatemodel}). Next, we present the details of these two interleaved processes.

Procedure~\ref{alg:framework} includes the steps of the learning process.
When the robot is free, it interacts with the environment by heuristically selecting a task\footnote{Here curriculum learning~\cite{narvekar2017autonomous} can play a role to task selection and we leave this aspect of the problem for future work.}, and repeatedly using a model-based RL approach, R-Max~\cite{brafman2002r} in our case, to complete the task.
The two guidance functions come from human knowledge.
For instance, given a navigation task, it comes from human knowledge that the robot should model its own position (specified by $f^V$) and actions that help the robot move between positions (specified by $f^A$).
After the policy converges or this learning process is interrupted (e.g., by task arrivals), the robot uses the learned probabilities to update the corresponding world dynamics in KRR.
For instance, the robot may have learned the probability and cost of navigating through a particular area in early morning.
In case this learning process is interrupted, the so-far-``known'' probabilities are used for knowledge base update.

% With our KRR-RL framework, the initial model of RL can be generated by KRR and the learned model of RL can update KRR in return, which potentially integrate reasoning and learning into an automated framework and decrease computational complexity of RL by using only task-relevant variables.

% Procedure~\ref{alg:framework} shows the main steps of our KRR-RL framework.
% For a certain task, it first differentiates all variables into endogenous and exogenous ones (Line 1), both which are used to generate transition probabilities (See Algorithm~\ref{alg:generatemodel}) as an initialized transition probabilities for an agent (Line 2-3). After the initialization, an agent uses a model-based RL algorithm (R-Max) to learn to complete the task (Line 4-5). At last, the new learned transition probabilities from RL are used to update old ones (Line 6).

Procedure~\ref{alg:generatemodel} includes the steps for building the probabilistic transition system of MDPs.
The key point is that we consider only endogenous variables in the task-specific state space.
However, when reasoning to compute the transition probabilities (Line~\ref{line:m}), the KRR component uses both $\Pi^P$ and $\mathcal{V}^{ex}$.
The computed probabilistic transition systems are used for building task-oriented controllers, i.e., $\pi$, for task completions.
In this way, the dynamically constructed controllers do not directly include exogenous variables, but their parameters already account for the values of all variables.

Next, we demonstrate how our KRR-RL framework is instantiated on a real robot.

%%%%%%%%%%%%%%%%%%%%%%%%%%%%%%%%%%%%%%%%%%%%%%%%%%%%%%%%%%
%%%%%%%%%%%%%%%%%%%%%%%%%%%%%%%%%%%%%%%%%%%%%%%%%%%%%%%%%%
%%%%%%%%%%%%%%%%%%%%%%%%%%%%%%%%%%%%%%%%%%%%%%%%%%%%%%%%%%

\subsection{An Instantiation on a Service Robot}
\label{sec:ins}
We consider a mobile service robot domain where a robot can do navigation, dialog, and delivery tasks.
A \emph{navigation task} requires the robot to use a sequence of (unreliable) navigation actions to move from one point to another.
In a \emph{dialog task}, the robot uses spoken dialog actions to specify service requests from people under imperfect language understanding.
A \emph{delivery task} requires the robot to use dialog to figure out the delivery request and conduct navigation tasks to physically fulfill the request.
Specifically, a delivery task requires the robot to deliver item \c{I} to room \c{R} for person \c{P}, resulting in services in the form of \c{<I,R,P>}.
The challenges come from unreliable human language understanding (e.g., speech recognition) and unforeseen obstacles that probabilistically block the robot in navigation.

Next, we present complete world models constructed using P-log~\cite{baral2009probabilistic,balai2017refining}, a logical-probabilistic KRR paradigm, and describe the process of constructing task-oriented controllers.

\begin{algorithm}[tb]\small
  \caption{Model Construction for Task Completion}
  \label{alg:generatemodel}
  \begin{algorithmic}[1]
    \REQUIRE{$\Pi^L$; $\Pi^P$; $\mathcal{V}^{en}$; $\mathcal{V}^{ex}$; Action set $A$}
%     \STATE{$f^A(\tau) \rightarrow A$}
    \FOR{$V_i \in \mathcal{V}^{en}$, $i~$ in $[0,\cdots,|\mathcal{V}^{en}|\!-\!1]$}
        \FOR{each possible value $v$ in $range(V_i)$}
            \FOR{each $a\in A$}
                \FOR{each possible value $v'$ in $range(V_i)$}
                    \STATE{$\mathcal{M}(v'|a,v) \leftarrow$ Reason with $\Pi^L$ and $\Pi^P$} w.r.t $\mathcal{V}^{ex}$ \label{line:m}
                \ENDFOR
            \ENDFOR
        \ENDFOR
    \ENDFOR
%     \STATE{Compute an action policy $\pi$ using $\mathcal{M}(v'|a,v)$}
    \RETURN $\mathcal{M}$
  \end{algorithmic}
\end{algorithm}

\begin{figure*}[tb]
\vspace{-1em}
  \begin{center}
    \includegraphics[width=0.8\textwidth]{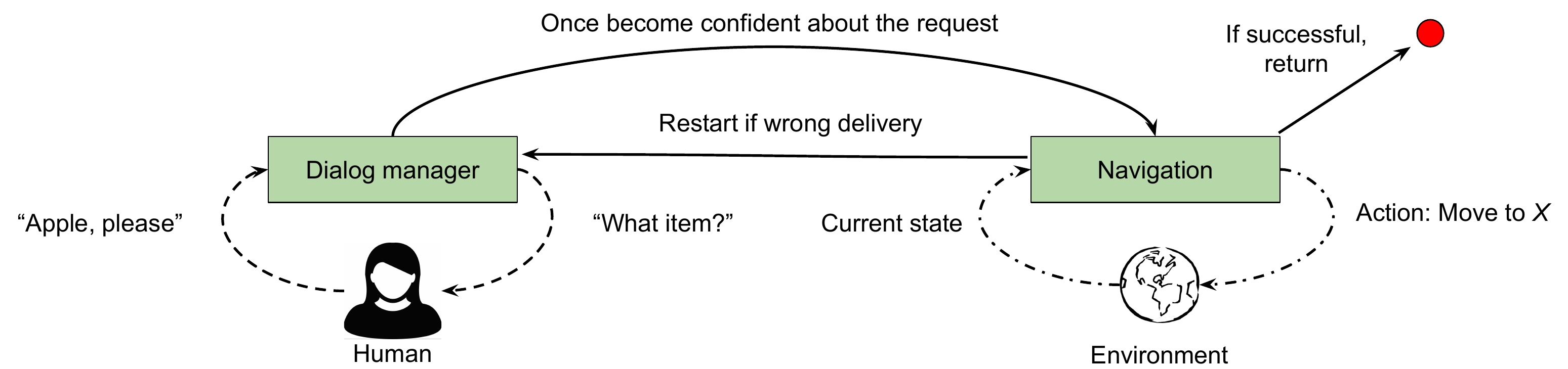}
  \end{center}
  \vspace{-1.5em}
  \caption{Transition system specified for delivery tasks, where question-asking actions are used for estimating the service request in dialog. Once the robot becomes confident about the service request, it starts to work on the navigation subtask. After the robot arrives, the robot might have to come back to the dialog subtask and redeliver, depending on whether the service request was correctly identified. }
  \label{fig:transition}
  \vspace{-1.5em}
\end{figure*}

%%%%%%%%%%%%%%%%%%%%%%%%%%%%%%%%%%%%%%%%%%%%%%%%%%%%%%%%%%
%%%%%%%%%%%%%%%%%%%%%%%%%%%%%%%%%%%%%%%%%%%%%%%%%%%%%%%%%%
% \vspace{-1em}
\paragraph{Representing Rigid Knowledge}

Rigid knowledge includes information that does not depend upon the passage of time. We introduce a set of \emph{sorts} including \c{time}, \c{person}, \c{item}, and \c{room}, so as to specify a complete space of service requests. For instance, we can use \c{time=\{morning, noon, afternoon, ...\}} to specify possible values of \c{time}.

% \begin{quote}
% \begin{scriptsize}
% \begin{verbatim}
% time={morning, noon, afternoon, ...}.
% person={alice, bob, carol, ...}.
% item={sandwich, coffee, ...}.
% room={office1, office2, lab, conf, ...}.
% boolean={true, false}.
% \end{verbatim}
% \end{scriptsize}
% \end{quote}

% We introduce more predicates to state that Alice is a professor, \c{prof\{alice\}}\footnote{P-log rule ends with a period punctuation. For the sake of readability, in this paper, it is omitted when a rule is embedded in text.},
% and Bob is a student, \c{student\{bob\}}. We can then specify Alice's office using \c{place(alice,office1)}, and state that students work in the lab using the rule of \c{place(P, lab):-student(P)}.

To model navigation domains, we introduce $N$ grid cells using \c{cell=\{0,...,N-1\}}, where the robot can travel. The geometric organization of these cells can be specified using predicates of \c{leftof($n_1$,$n_2$)} and \c{belowof($m_1$,$m_2$)}, where $n_i$ and $m_j$ are cells, $i,j\in \{0,\dots, N-1\}$.

We use a set of random variables to model the space of world states.
For instance, \c{curr\_cell:cell} states that the robot's position \c{curr\_cell}, as a random variable, must be in sort \c{cell};
\c{curr\_room:person->room} states that a person's current room must be in sort \c{room};
and \c{curr\_succ} identifies the request being fulfilled or not.
%
% \begin{quote}
% \begin{scriptsize}
% \begin{verbatim}
% curr_cell:cell.           curr_item:person->item.
% curr_time:time.           curr_room:person->room.
% curr_person:person.       curr_succ:boolean.
% \end{verbatim}
% \end{scriptsize}
% \end{quote}
%
We can then use \c{random(curr\_cell)} to state that the robot's current position follows a uniform distribution over \c{cell}, unless specified elsewhere; and use the following pr-atom (a form of declarative probabilistic rule in P-log) to state that, in probability 0.8, people request deliveries to their own rooms.

\begin{quote}
\begin{scriptsize}
\begin{verbatim}
pr(curr_room(P)=R|place(P,R)=true)=8/10.
\end{verbatim}
\end{scriptsize}
\end{quote}

%%%%%%%%%%%%%%%%%%%%%%%%%%%%%%%%%%%%%%%%%%%%%%%%%%%%%%%%%%
%%%%%%%%%%%%%%%%%%%%%%%%%%%%%%%%%%%%%%%%%%%%%%%%%%%%%%%%%%
% \vspace{-1em}
\paragraph{Representing Action Knowledge}

We use a set of random variables in P-log to specify the set of delivery actions available to the robot. For instance, \c{serve(coffee,lab,alice)} indicates that the current service request is to deliver \c{coffee} to \c{lab} for \c{alice}.
\begin{quote}
\begin{scriptsize}
\begin{verbatim}
serve(I,R,P) :- act_item=I, act_room=R, act_person=P.
\end{verbatim}
\end{scriptsize}
\end{quote}
where predicates with prefix \c{act\_} are random variables for modeling \c{serve} actions.

To model probabilistic transitions, we need to look one step forward, modeling how actions lead state transitions. We introduce two identical state spaces using predicates \c{curr\_s} and \c{next\_s}. The following shows the specification of the current state using \c{curr\_s}.

\begin{quote}
\begin{scriptsize}
\begin{verbatim}
curr_s(I,R,P,C,S) :- curr_item(P)=I, curr_room(P)=R,
            curr_person=P, curr_cell=C, curr_succ=S.
\end{verbatim}
\end{scriptsize}
\end{quote}

% \begin{quote}
% \begin{scriptsize}
% \begin{verbatim}
% next_s(I,R,P,C,S) :- next_item(P)=I, next_room(P)=R,
%                      next_person=P, next_cell=C,
%                      next_succ=S.                (R3)
% \end{verbatim}
% \end{scriptsize}
% \end{quote}

% The following are two example pr-atoms that specify the transition probabilities led by delivery actions. The first specifies the probability of successfully accomplishing a delivery task, when the requested delivery is correctly identified:
% \begin{quote}
% \begin{scriptsize}
% \begin{verbatim}
% pr(next_succ=true | curr_s(I,P,R,C,false),
%                     serve(I,P,R) = 9/10.         (R4)
% \end{verbatim}
% \end{scriptsize}
% \end{quote}
% and the second specifies the probability, when the item component of the requested delivery is incorrectly identified:
% \begin{quote}
% \begin{scriptsize}
% \begin{verbatim}
% pr(next_succ=true | curr_s(IS,PS,RS,CS,false),
%                     serve(I,P,R), IS<>I) = 3/20. (R5)
% \end{verbatim}
% \end{scriptsize}
% \end{quote}

Given the current and next state spaces, we can use pr-atoms to specify the transition probabilities led by delivery actions.
For instance, to model physical movements, we introduce random variable \c{act\_move:move} and sort \c{move=\{up, down, left, right\}}. Then we can use \c{act\_move=right} to indicate the robot attempting to move rightward by one cell, and use the following pr-atom
\begin{quote}
\begin{scriptsize}
\begin{verbatim}
pr(next_cell=C1 | curr_cell=C, leftof(C,C1),
                  act_move=right) = 8/10.
\end{verbatim}
\end{scriptsize}
\end{quote}
to state that, after taking action \c{right}, the probability of the robot successfully navigating to the cell on the right is $0.8$ (otherwise, it ends up with one of the nearby cells).
Such probabilities are learned through model-based RL.

It should be noted that, even if a request is correctly identified in dialog, the robot still cannot always succeed in delivery, because there are obstacles that can probabilistically trap the robot in navigation.
%Success rates of such deliveries are modeled using rules such as \c{(R4)}.
When the request is misidentified, delivery success rate drops, because the robot has to conduct multiple navigation tasks to figure out the correct request and redo the delivery.
We use $s\odot a$ and $s\otimes a$ to represent delivery action $a$ matches to request component of $s$ or not (i.e., service request is correctly identified in dialog or not).

% \begin{figure}[tb]\vspace{-.2em}
%   \begin{center}
%     \includegraphics[width=0.48\textwidth]{images/trans_nav}
%   \end{center}
%   \vspace{-.5em}
%   \caption{Some text. Some text. Some text. Some text. Some text. Some text. Some text. Some text. Some text. Some text. Some text. Some text. Some text. Some text. Some text. }
%   \label{fig:trans_nav}
%   %\vspace{-.5em}
% \end{figure}

% Figure~\ref{fig:trans_nav} shows an example of a navigation task, where the state space includes only two states.

%%%%%%%%%%%%%%%%%%%%%%%%%%%%%%%%%%%%%%%%%%%%%%%%%%%%%%%%%%
%%%%%%%%%%%%%%%%%%%%%%%%%%%%%%%%%%%%%%%%%%%%%%%%%%%%%%%%%%
% \vspace{-1em}
\paragraph{Constructing (PO)MDP Controllers}

To fulfill a delivery request, the robot needs spoken dialog to identify the request under unreliable speech recognition, and navigation controllers for physically making the delivery.

The service request is not directly observable to the robot, and has to be estimated by asking questions, such as ``What item do you want?'' and ``Is this delivery for Alice?''
Once the robot is confident about the request, it takes a delivery action (i.e., \c{serve(I,R,P)}).
We follow a standard way to use partially observable MDPs (POMDPs)~\cite{kaelbling1998planning} to build our dialog manager, as reviewed in~\cite{Young2013POMDP}.
The state set $\mathcal{S}$ is specified using \c{curr\_s}.
The action set $\mathcal{A}$ is specified using \c{serve} and question-asking actions.
Question-asking actions do not change the current state, and delivery actions lead to one of the terminal states (success or failure).

After the robot becomes confident about the request via dialog, it will take a delivery action \c{serve\{I,R,P\}}. This delivery action is then implemented with a sequence of \c{act\_move} actions. When the request identification is incorrect, the robot needs to come back to the shop, figure out the correct request, and redeliver, where we assume the robot will correctly identify the request in the second dialog.
We use an MDP to model this robot navigation task, where the states and actions are specified using sorts \c{cell} and \c{move}. We use pr-atoms to define the unreliable movements.
Figure~\ref{fig:transition} shows the probabilistic transitions in delivery tasks.

%%%%%%%%%%%%%%%%%%%%%%%%%%%%%%%%%%%%%%%%%%%%%%%%%%%%%%%%%%
%%%%%%%%%%%%%%%%%%%%%%%%%%%%%%%%%%%%%%%%%%%%%%%%%%%%%%%%%%
% \vspace{-1em}
\paragraph{Knowledge from Model-based RL}

We use R-Max~\cite{brafman2002r}, a model-based RL algorithm, to help our robot learn the success rate of navigation actions in different positions.
% We first initialize an MDP $\langle \mathcal{S}, \mathcal{A}, T^N, \mathcal{R}\rangle$, from which we use R-Max to learn the partial world model.
% $\mathcal{S}$ includes a virtual state $s^v$ and a set of $N+1$ states $\{s_0,\cdots,s_N\}$ that correspond to the set of cells in our reasoner specified using \c{cell=\{0..N\}} in P-log.
% $\mathcal{A}$ is the action set that includes four actions in the sort of \c{move} as described in Section~\ref{sec:action}.
The agent first initializes an MDP, from which it uses R-Max to learn the partial world model (of navigation tasks).
% Specifically, it initializes the transition function with $T^N(s,a,s^v)=1.0$, where $s\in \mathcal{S}$ and $a\in \mathcal{A}$, meaning that starting from any state, after any action, the next state is always $s^v$. The reward function is initialized with $\mathcal{R}(s,a)=R_{max}$, where $R_{max}$ is an upper bound of reward. The initialization of $T^N$ and $\mathcal{R}$ enables the learner to automatically balance exploration and exploitation.
There is a fixed small cost for each navigation action. The robot receives a big bonus if it successfully achieves the goal ($R_{max}$), whereas it receives a big penalty otherwise ($-R_{max}$).
%Once robot select an action $a$ at a state $s$, we give a fixed reward -1, and if the robot reach the target office in limited time steps, it will receive the maximum reward $R_{max}=40$, or the minimum reward -$R_{max}$ received.
A transition probability in navigation, $T^N(s,a,s')$, is not computed until there are a minimum number ($M$) of transition samples visiting $s'$.
% We follow the R-Max paper to initialize $M=0$, but it can be an arbitrary number in practice.
We recompute the action policy after $E$ action steps.

The update of knowledge base is achieved through updating the success rate of delivery actions \c{serve(I,R,P)} (in dialog task) using the success rate of navigation actions \c{act\_move=M} in different positions (in navigation task).
\begin{align}
\label{eqn:transfer}
    &T^D(s^r, a^d, s^t)= \nonumber\\
    &\begin{cases}
        P^N(s^{sp}\!, s^{gl}),~{\bf if}~s^r\odot a^d\\
        P^N(s^{sp}\!, s^{mi})\times P^N(s^{mi}\!, s^{sp}) \times P^N(s^{sp}\!, s^{gl}),~{\bf if}~s^r\otimes a^d
    \end{cases}
\end{align}
where $T^D(s^r, a^d, s^t)$ is the probability of fulfilling request $s^r$ using delivery action $a^d$;
$s^t$ is the ``success'' terminal state;
$s^{sp}$, $s^{mi}$ and $s^{gl}$ are states of the robot being in the shop, a misidentified goal position, and real goal position respectively;
and $P^N(s,s')$ is the probability of the robot successfully navigating from $s$ to $s'$ positions. In practice, $P^N(s,s')$ is approximated by running a number of navigation tasks in simulation using the learned partial world model, while following an action policy computed using value iteration.

When $s^r$ and $a^d$ are aligned in all three dimensions (i.e., $s^r\odot a^d$), the robot needs to navigate once from the shop ($s^{sp}$) to the requested navigation goal ($s^{gl}$).
$P^N(s^{sp},s^{gl})$ is the probability of the corresponding navigation task. When the request and delivery action are not aligned in at least one dimension (i.e., $s^{r}\otimes a^{d}$), the robot has to navigate back to the shop to figure out the correct request, and then redeliver, resulting in three navigation tasks.

\section{Experiments}
\label{sec:experiment}

\begin{figure}[tb]
 \vspace{.5em}
  \begin{center}
    \includegraphics[height=0.15\textheight]{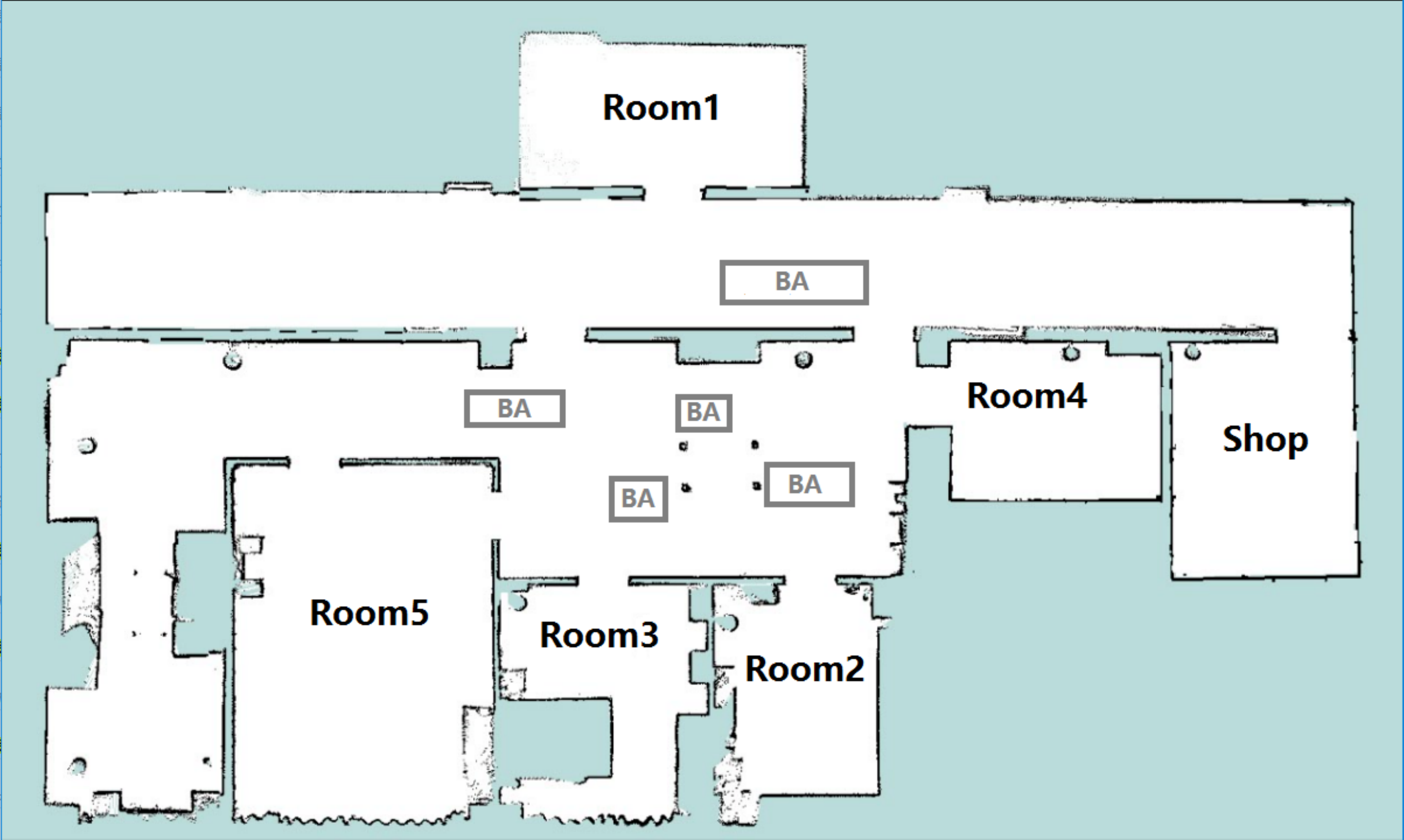}
    \includegraphics[height=0.15\textheight]{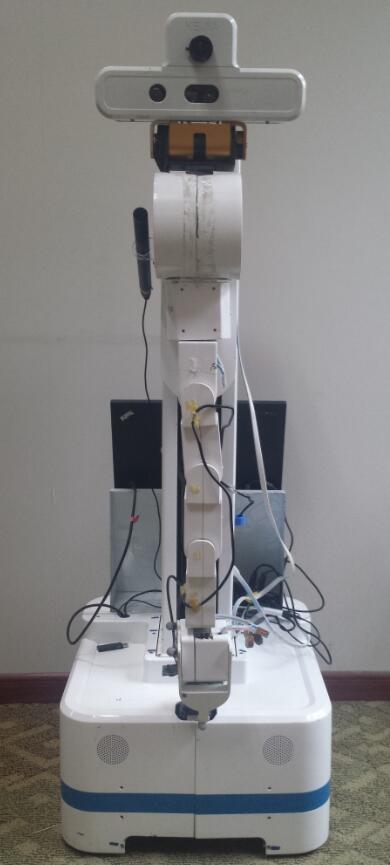}
  \end{center}
  \vspace{-.5em}
  \caption{Occupancy-grid map used in our experiments, including five rooms, one shop, and four blocking areas (indicated by `BA'). All deliveries are from the shop and to one of the rooms. }
  \label{fig:map}
  \vspace{-.5em}
\end{figure}

% We evaluate our guided TL-for-RL framework using a mobile robot that works in an office environment. The robot needs to use spoken dialog to figure out human requests, and then physically conduct the delivery $\langle I,P,R\rangle$. When there is no task assigned to the robot, it randomly selects a navigation goal, from which it learns the partial world model related to navigation tasks. Experiments have been conducted comprehensively in simulation  using Gazebo, a simulation environment with physics engine~\cite{Koenig04designand}, and the algorithms have been implemented and demonstrated using a real robot. Figure~\ref{fig:map} shows the domain map used in all experiments.

In this section, the goal is to evaluate our hypothesis that our KRR-RL framework enables a robot to learn from model-based RL, reason with both the learned knowledge and human knowledge, and dynamically construct task-oriented controllers.
Specifically, our robot learns from navigation tasks, and applied the learned knowledge (through KRR) to navigation, dialog, and delivery tasks.
We also evaluated whether the learned knowledge can be represented and applied to tasks under different world settings.
In addition to simulation experiments, we have used a real robot to demonstrate how our robot learns from navigation to perform better in dialog.
Figure~\ref{fig:map} shows the map of the working environment (generated using a real robot) used in both simulation and real-robot experiments. Human walkers in the blocking areas (``BA'') can probabilistically impede the robot, resulting in different success rates in navigation tasks.

% Some human walker models are added in order to simulate the blocking areas and through adjusting the number or the moving frequency of human walkers we can simulate the different blocking rate (the higher of blocking rate, the harder for the robot to pass).

% \begin{figure}[tb]
%   \vspace{.5em}
%   \begin{center}
%     \includegraphics[width=0.52\textwidth]{images/blocking_rate.png}
%   \end{center}
%   \vspace{-.5em}
%   \caption{Navigation success rate on different blocking rate.}
%   \label{fig:blockingrate}
%   \vspace{-.5em}
% \end{figure}

% \comments{This figure seems unnecessary to me, which means we can support our hypothesis without using this figure. As long as the domain map is clearly marked, the results in this figure do not produce any new information. } Figure~\ref{fig:blockingrate} shows the impact of blocking rate on navigation to each office. We use the 0.3 as a default blocking rate in all experiments unless explicitly stated. We split the world into many grid cells with different size, and adjust the size of grid cells to restrict the total number (nearly 100 grid cells). The algorithm and its detail parameters used in navigation tasks have been introduced in Section~\ref{sec:learn}.

% When the blocking rate is 0.3 for all trapping areas, the navigation success rates to offices 2 and 5 are 0.643 and 0.519. When the blocking rate is increased to 0.7, the success rates drop to 0.237 and 0.11. The success rate to office 1 is close to 1.0 and does not change significantly.

\begin{figure}[tb]
%   \vspace{-1.5em}
  \begin{center}
    \hspace*{-.1em}\includegraphics[width=0.26\textwidth]{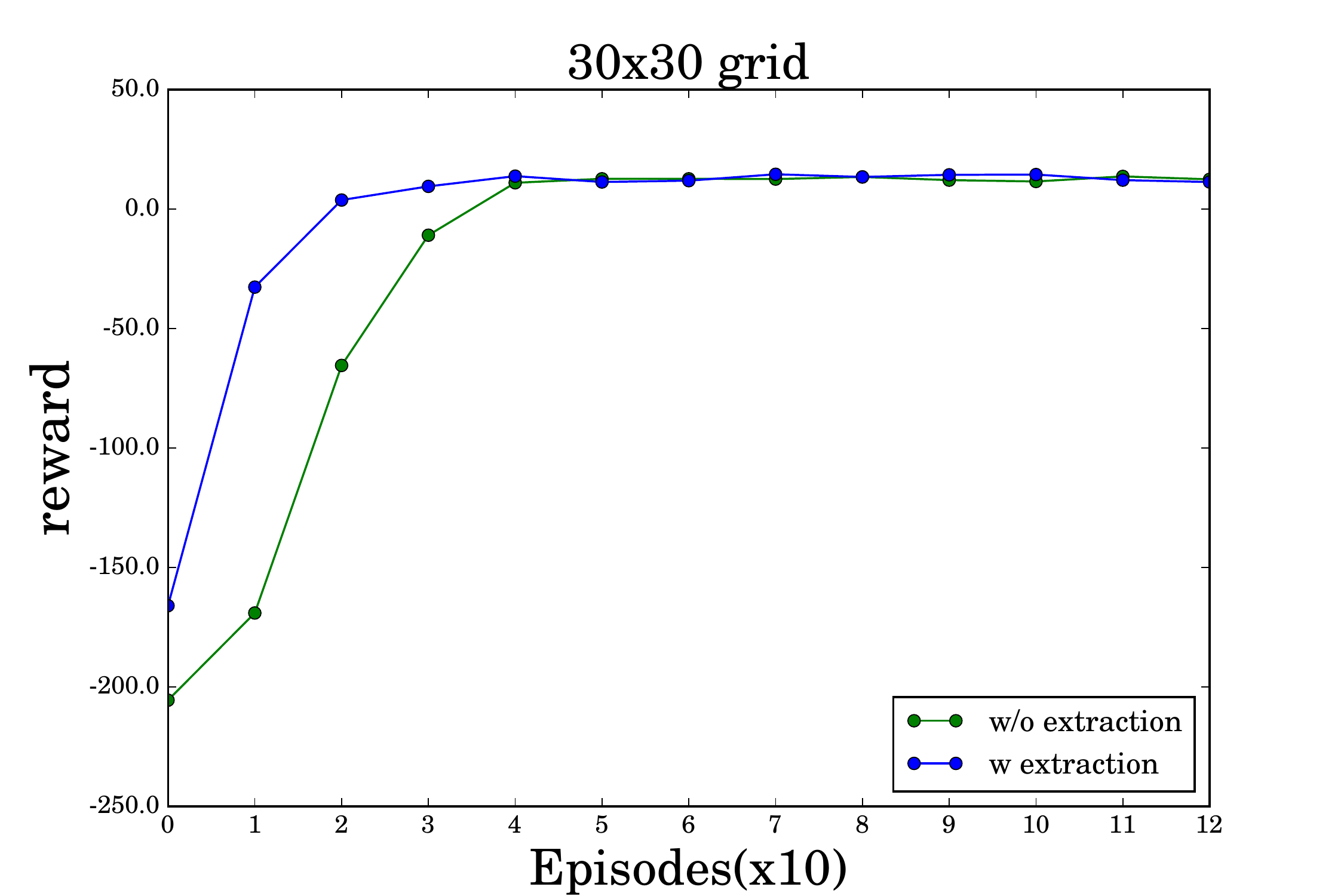} \hspace*{-1.7em}
    \includegraphics[width=0.26\textwidth]{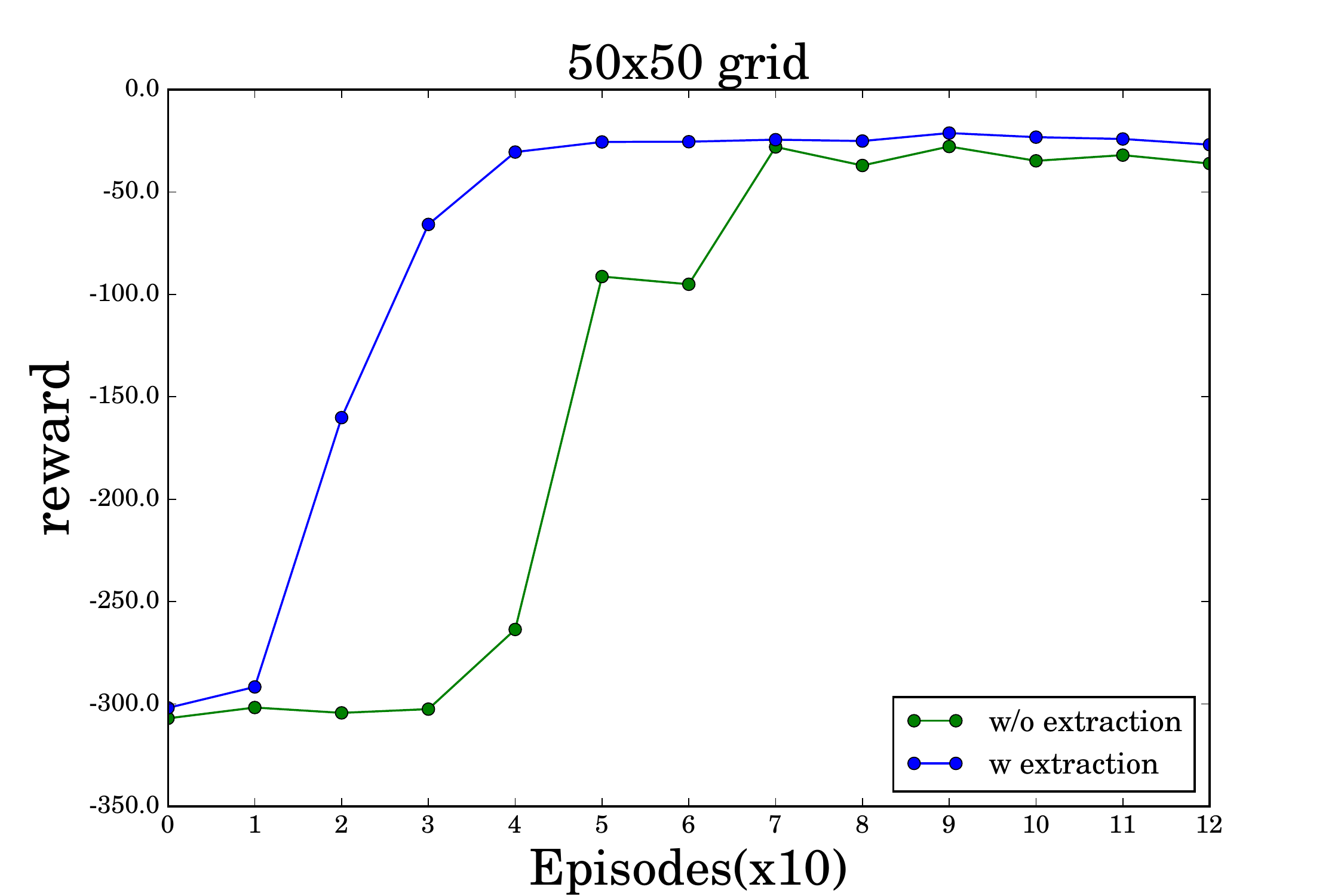}\hspace*{-1.4em}
  \end{center}
  \vspace{-1.5em}
  \caption{Navigation tasks in small (Left: $30\times 30$ grid) and large (Right: $50\times 50$ grid) domains.
  With extraction (dashed line), the robot learns faster in the target navigation task. }
  \label{fig:navigation}
  \vspace{-1.5em}
\end{figure}

% \vspace{-1em}
\paragraph{Learn from and applied to navigation tasks}
Focusing on navigation tasks, in this experiment, the robot learns in the \c{shop-room1} navigation task, and extracts the learned partial world model to the \c{shop-room2} task. It should be noted that navigation from \c{shop} to \c{room2} requires traveling in areas that are unnecessary in the \c{shop-room1} task. The results are shown in Figure~\ref{fig:navigation}, where each data points corresponds to an average of 1000 trials. Each episode allows at most 200 (300) steps in small (large) domain. The curves are smoothed using a window of 10 episodes. The results suggest that with knowledge extraction (the dashed line) the robot learns faster than without extraction, and this performance improvement is more significant in a larger domain (the Right subfigure).

% \vspace{-1em}
\paragraph{Learn from navigation and applied to delivery}

Robot delivering objects requires both tasks: dialog management for specifying service request (under unreliable speech recognition) and navigation for physically delivering objects (under unforeseen obstacles).
Our office domain includes five rooms, two persons, and three items, resulting in 30 possible service requests.
In the dialog manager, the reward function gives delivery actions
\begin{itemize}
\item  a big bonus (80) given a request being fulfilled, and
\item  a big penalty (-80) otherwise.
\end{itemize}
General questions and confirming questions cost 2.0 and 1.5 respectively. In case a dialog does not end after 20 turns, the robot is forced to work on the most likely delivery.
%We model speech recognition errors (e.g. 0.8 accuracy in recognizing ``yes/no'').

\renewcommand{\arraystretch}{1}
\begin{table}[t]
  \centering
  \fontsize{6.5}{8}\selectfont
  \begin{threeparttable}
  \caption{Overall performance of both question-asking in dialog management and navigation in object delivery. }
  \label{tab:both}
    \begin{tabular}{ccccccc}
    \toprule
    \multirow{2}{*}{}&\multicolumn{3}{c}{Without extraction}&\multicolumn{3}{c}{With extraction}\cr
    \cmidrule(lr){2-4} \cmidrule(lr){5-7}
    & Reward & Fulfilled & QA Cost & Reward & Fulfilled & QA Cost \cr
    \midrule
    $br=0.1$ &182.07 &0.851 &20.86     &206.21 &0.932 &18.73 \cr
    $br=0.5$ & 30.54 &0.853 &20.84     & 58.44 &0.927 &18.98 \cr
    $br=0.7$ &-40.33 &0.847 &20.94     &-14.50 &0.905 &20.56 \cr
    \bottomrule
    \end{tabular}
      \vspace{-1.5em}
    \end{threeparttable}
\end{table}

Table~\ref{tab:both} reports the robot's overall performance in fulfilling delivery requests, which requires the robot accurately identifying the request in dialog and then safely delivering the item in navigation.
We conduct 10,000 simulation trials under each blocking rate.
Without learning from RL, the robot uses an outdated world model that was learned under $br=0.3$. With learning, the robot updates its world model in domains with different blocking rates.
We can see, when learning is enabled, our KRR-RL framework produces higher overall reward, higher request fulfillment rate, and lower question-asking cost.
The improvement is statistically significant, e.g., the $p\!-\!values$ are 0.028, 0.035, and 0.049 for overall reward, when \emph{br} is 0.1, 0.5, and 0.7 respectively (100 randomly selected trials with/without extraction).

\begin{figure}[thb]
  \vspace{-1em}
  \begin{center}
    \includegraphics[width=0.26\textwidth]{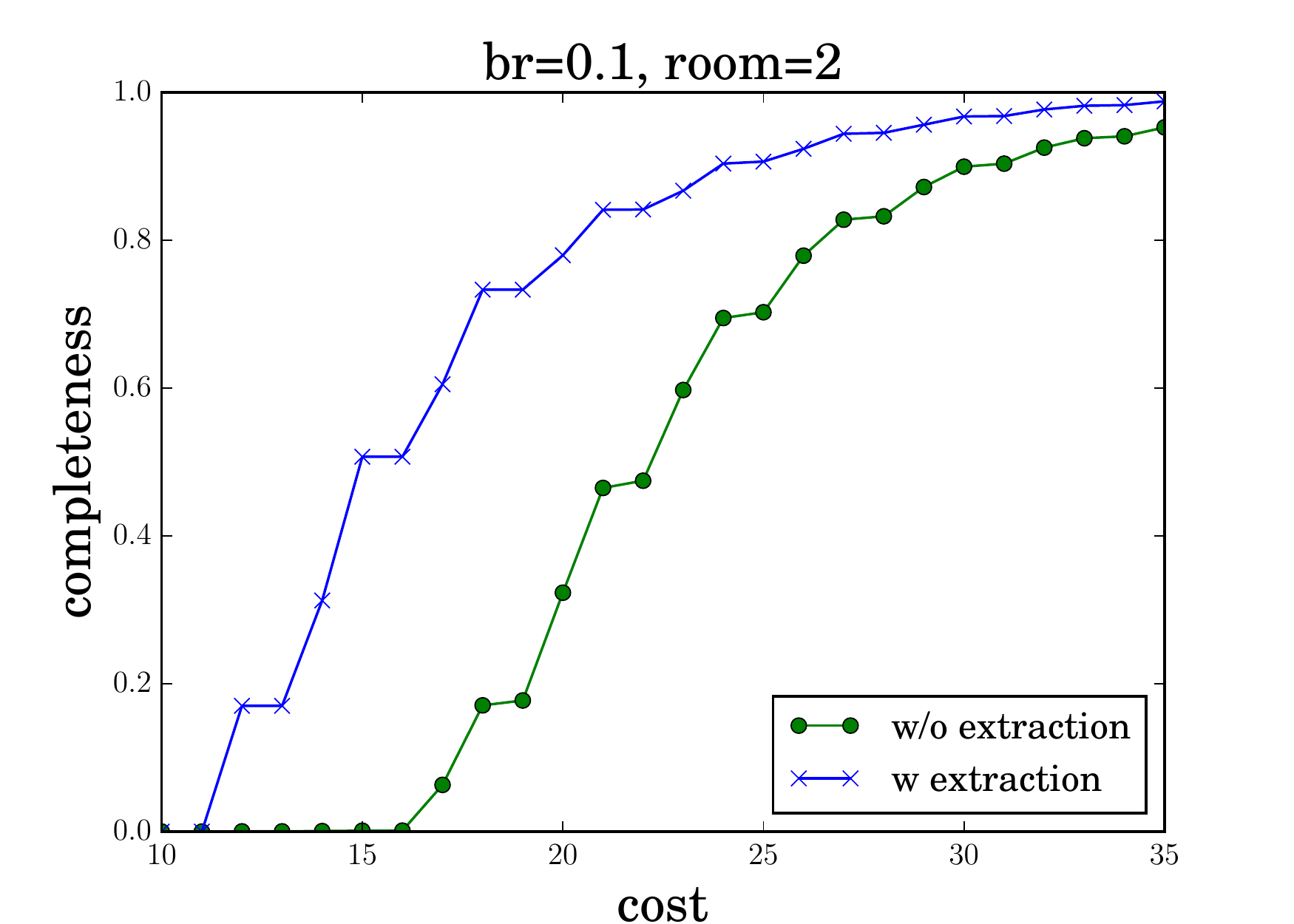}\hspace*{-1.5em}
    \includegraphics[width=0.26\textwidth]{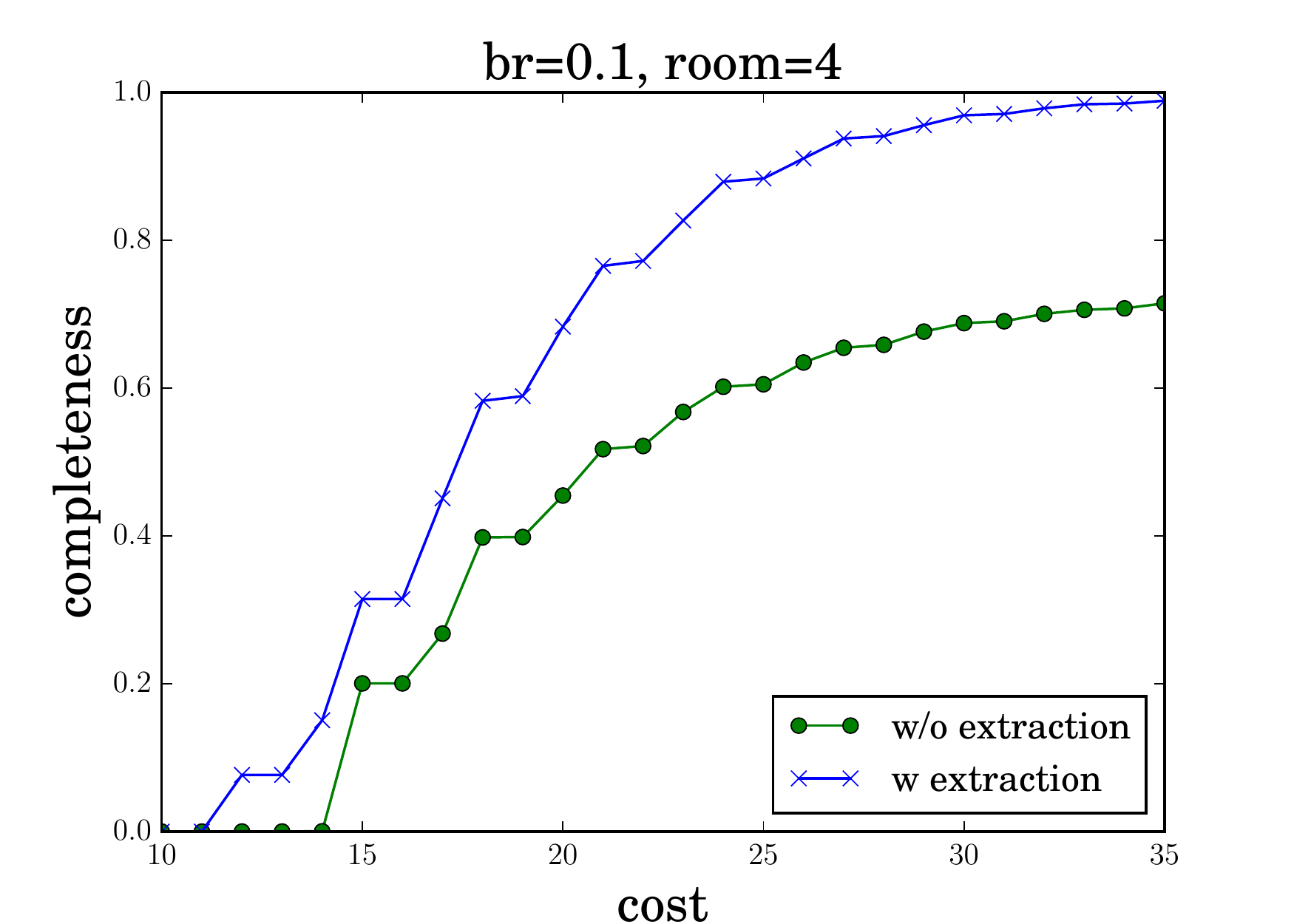}\hspace*{-1em}\vspace{-.5em}

  \end{center}
  \vspace{-1em}
  \caption{CDF plots of dialog completions. The requests are deliveries to \c{room2} (Left) and \c{room4} (right). }
  \label{fig:completeness}
  \vspace{-.5em}
\end{figure}

To better analyze robot dialog behaviors, we generate cumulative distribution function (CDF) plots showing the percentage of dialog completions (y-axis) given different QA costs (x-axis). Figure~\ref{fig:completeness} shows the results when the request are deliveries to \c{room2} (left) and \c{room4} (right).
Comparing the two curves in each subfigure, we find our KRR-RL framework reduces the QA cost in dialogs (consistent to Table~\ref{tab:both}).
% Comparing the left (or right) two subfigures, we find a lower dialog completion rate given a higher $br$.
%For instance, when the QA cost is 20, the percentage of dialog completion is reduced from 0.8 to 0.6 when \emph{br} grows from 0.1 to 0.5.
% This indicates that our dialog manager becomes more conservative given higher blocking rates, which meets our expectation.
Comparing the two subfigures, we find a smaller QA cost is needed, when the request is a delivery to \c{room4} (c.f., \c{room2}). This observation makes sense, because \c{room4} is closer to the shop (see Figure~\ref{fig:map}) and deliveries to \c{room4} is easier.

In the last experiment, we quantify the information collected in dialog in terms of entropy reduction. The hypothesis is that, using our KRR-RL framework, the dialog manager wants to collect more information before physically working on more challenging tasks.
In each trial, we randomly generate a belief distribution over all possible service requests, evaluate the entropy of this belief, and record the suggested action given this belief. We then statistically analyze the entropy values of beliefs, under which delivery actions are suggested.

Table~\ref{tab:entropy} shows that, when \emph{br} grows from 0.1 to 0.7, the means of belief entropy decreases (i.e., belief is more converged).
This suggests that the robot collected more information in dialog in environments that are more challenging for navigation, which is consistent with Table~\ref{tab:both} (Right).
Comparing the three columns of results, we find the robot collects the most information before it delivers to \c{room5}.
This is because such delivery tasks are the most difficult due to the location of \c{room5}. The results support our hypothesis that learning from navigation tasks enables the robot to adjust its information gathering strategy in dialog given tasks of different difficulties.

\renewcommand{\arraystretch}{1}
\begin{table}[t]
  \centering
  \fontsize{6.5}{8}\selectfont
   \vspace{.5em}
  \begin{threeparttable}
  \caption{The amount of information (in terms of entropy) needed by a robot before taking delivery actions.}
  \vspace{-.5em}
  \label{tab:entropy}
    \begin{tabular}{ccccccc}
    \toprule
    \multirow{3}{*}{}   &\multicolumn{2}{c}{Entropy (room1)}
                        &\multicolumn{2}{c}{Entropy (room2)}
                        &\multicolumn{2}{c}{Entropy (room5)}\cr
    \cmidrule(lr){2-3} \cmidrule(lr){4-5} \cmidrule(lr){6-7}
    & Mean (std) & Max & Mean (std) & Max & Mean (std) & Max \cr
    \midrule
    $br=0.1$ & .274 (.090) & .419 & .221 (.075)     & .334 & .177 (.063) & .269 \cr
    $br=0.5$ & .154 (.056) & .233 & .111 (.044)     & .176 & .100 (.041) & .156 \cr
    $br=0.7$ & .132 (.050) & .207 & .104 (.042)     & .166 & .100 (.041) & .156 \cr
    \bottomrule
    \end{tabular}
    \vspace{-2em}
    \end{threeparttable}
\end{table}

\paragraph{Generate controllers for new circumstances}

The knowledge learned through model-based RL is contributed to a knowledge base that can be used for many tasks.
So our KRR-RL framework enables a robot to dynamically generate partial world models for tasks under settings that were never experienced.
For example, an agent does not know the current time is morning or noon, there are two possible values for variable ``time''.
Consider that our agent has learned world dynamics under the times of morning and noon.
Our KRR-RL framework enables the robot to reason about the two transition systems under the two settings and generate a new transition system for this ``morning-or-noon'' setting.
Without our framework, an agent would have to randomly select one between the ``morning'' and ``noon'' policies.

To evaluate our policies dynamically constructed via KRR, we let an agent learn three controllers under three different environment settings -- the navigation actions have decreasing success rates under the settings.
In this experiment, the robot does not know which setting it is in (out of two that are randomly selected).
The baseline agent does not has the KRR capability of merging knowledge learned from different settings, and can only randomly select a policy from the two (each corresponding to a setting).
Experimental results show that the baseline agent received an average of $26.8\%$ in success rate in navigation tasks.
In comparison, our KRR-RL agent achieved $83.8\%$ success rate on average.
Thus, our KRR-RL framework enables a robot to effectively apply the learned knowledge to tasks under new settings.

% \vspace{-1em}
\paragraph{An Illustrative Trial on a Robot}
%We have also implemented our approach on KeJia robot~\cite{Chen2011KeJia} in a office environment shown in Figure~\ref{fig:KeJia}. It is equipped with two Lidar sensors, a Kinect RGB-D camera for navigation. We use the Speech Application Programming Interface (SAPI) package\footnote{\url{http://www.iflytek.com/en}}. Based on these settings, we run the Robot Operating System on KeJia. Given two requests (one for office1 and another for office5) and the success rates of navigation, KeJia has two different dialog styles, where the former request takes less dialog turns before KeJia makes a delivery action. The demo video is available online.

% We have implemented our TL-for-RL framework on a KeJia robot in an office environment.
% As shown in Figure~\ref{fig:map} (Right)
% the robot is equipped with two Lidar sensors for localization and obstacle avoidance in navigation, and a Kinect RGB-D camera for human-robot interaction.
% We use the Speech Application Programming Interface (SAPI) package\footnote{\url{http://www.iflytek.com/en}} for speech recognition.
% The robot software runs in the Robot Operating System (ROS)~\cite{Quigley:icra09}.
% Before KeJia initiates a conversation, it collects facts from the environment and reasons about the facts to initialize its belief over possible service requests.

\begin{figure}[tb]
  \vspace{-1em}
  \hspace*{-.5em}
  \begin{center}
    \includegraphics[width=0.48\textwidth]{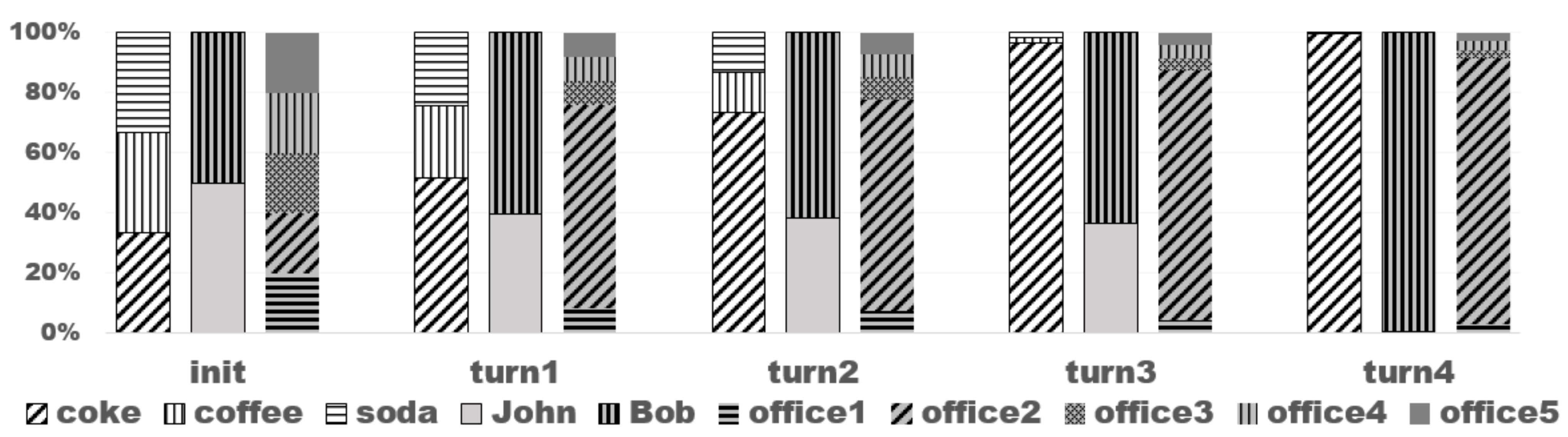}
  \end{center}
  \vspace{-1em}
  \caption{Belief change in three dimensions (In order from the left: Items, Persons and Offices) over five turns in a human-robot dialog .
  % The distributions are grouped by turns (Including the initial distribution). In each turn, there are three distribution bars which means three different dimensions (In order from the left: Item, Person and Office). In order from the bottom, the values in each dimension are 1) coke, coffee and soda in Item; 2) John and Bob in Person; and 3) office1, office2, office3, office4 and office5 in Office.
  }
  \label{fig:distribution}
  \vspace{-1.5em}
\end{figure}

We have implemented our KRR-RL framework on a mobile robot in an office environment, the robot is shown in Figure~\ref{fig:map} (Right).
% Table~\ref{dialog} shows the dialog between KeJia and a user using navigation knowledge transfer.
Figure~\ref{fig:distribution} shows the belief changes (in the dimensions of \c{item}, \c{person}, and \c{room}) as the robot interacts with a human user.
The robot started with a uniform distribution in all three categories. It should be noted that, although the marginal distributions are uniform, the joint belief distribution is not, as the robot has prior knowledge such as \c{Bob}'s office is \c{office2} and people prefer deliveries to their own offices.

After hearing ``a coke for Bob to office2'', the three sub-beliefs are updated (\c{turn1}).
Since the robot is aware of its unreliable speech recognition, it asked about the item, ``Which item is it?''
After hearing ``a coke'', the belief is updated (\c{turn2}), and the robot further confirmed on the item by asking ``Should I deliver a coke?''
It received a positive response (\c{turn3}), and decided to move on to ask about the delivery room: ``Should I deliver to office 2?''
\emph{The robot did not confirm the delivery room, because it learned through model-based RL that navigating to \c{office2} is relatively easy and it decided that it is more worth risking an error and having to replan than it is to ask the person another question. }
The robot became confident in three dimensions of the service request (\c{<coke,Bob,office2>} in \c{turn4}) \emph{without} asking about \c{person}, because of the prior knowledge (encoded in P-log) about \c{Bob}'s office.
% A demo video is available online.\footnote{https://youtu.be/SxhAcJHfkYI}

% \begin{table}\footnotesize
% \begin{center}
% \caption{Dialog Demo with Transfer Knowledge.}
% \label{dialog}
% \begin{tabular}{|l|}
% \hline
% User: I'd like to order a coke for Bob to office2.  \\
% Sys: I am sorry. Which item is it? \\
% User: A coke. \\
% Sys: Should I deliver a coke? \\
% User: Yes. \\
% Sys: Should I deliver to office2? \\
% User: Yes. \\
% Sys: Ok, I am delivering a coke for Bob to office2. \\\hline
% \end{tabular}
% \end{center}
% \end{table}

%%%%%%%%%%%%%%%%%%%%%%%%%%%%%%%%%%%%%%%%%%%%%%%%%%%%%%%%%%
%%%%%%%%%%%%%%%%%%%%%%%%%%%%%%%%%%%%%%%%%%%%%%%%%%%%%%%%%%
%%%%%%%%%%%%%%%%%%%%%%%%%%%%%%%%%%%%%%%%%%%%%%%%%%%%%%%%%%

\section{Conclusions and Future Work}
\label{sec:conclude}
In this work, we develop a KRR-RL framework that, for the first time, integrates computational paradigms of logical-probabilistic knowledge representation and reasoning (KRR), and model-based reinforcement learning (RL).
% The KRR component was realized using a logical-probabilistic declarative programming paradigm called P-log, and the RL component was realized using R-Max.
Our KRR-RL agent learns world dynamics (in the form of transition probabilities) via model-based RL, and then incorporates the learned dynamics into the logical-probabilistic reasoning module, which is used for dynamic construction of efficient run-time task-specific planning models.
% Transition probabilities that result from actions are learned via model-based RL, and then incorporated into the probabilistic reasoning module, which in turn enables dynamic construction of efficient run-time task-specific planning models.
Experiments were conducted using a mobile robot (simulated and physical) working on delivery tasks that involve both navigation and dialog.
Results suggested that the learned knowledge from RL can be represented and used for reasoning by the KRR component, enabling the robot to dynamically generate task-oriented action policies.

% We use a KRR formalism to maintain a complete world model, where its parameters can be updated by model-based RL, and its partial models are used for building probabilistic controllers.
% The framework has been evaluated using a delivery robot that extracts knowledge from navigation to dialog tasks.

The integration of a KRR paradigm and model-based RL paves the way for  at least the following research directions.
We plan to study how to sequence source tasks to help the robot perform the best in the target task (i.e., a curriculum learning problem~\cite{narvekar2017autonomous}).
%We also plan to evaluate the performance of transfer from dialog (where feedback is more sparse) to navigation.
Balancing the efficiencies between service task completion and RL is another topic for further study -- currently the robot optimizes for task completions (without considering the potential knowledge learned in this process) once a task becomes available.

{
\bibliographystyle{plain}
\bibliography{references}

\begin{thebibliography}{10}

\bibitem{asgharbeygi2006relational}
Nima Asgharbeygi, David Stracuzzi, and Pat Langley.
\newblock Relational temporal difference learning.
\newblock In {\em Proceedings of the 23rd international conference on Machine
  learning}, pages 49--56. ACM, 2006.

\bibitem{bach2017hinge}
Stephen~H. Bach, Matthias Broecheler, Bert Huang, and Lise Getoor.
\newblock Hinge-loss markov random fields and probabilistic soft logic.
\newblock {\em Journal of Machine Learning Research}, 18(1):3846--3912, 2017.

\bibitem{balai2017refining}
Evgenii Balai and Michael Gelfond.
\newblock Refining and generalizing p-log: Preliminary report.
\newblock In {\em Proceedings of the 10th Workshop on Answer Set Programming
  and Other Computing Paradigms}, 2017.

\bibitem{baral2009probabilistic}
Chitta Baral, Michael Gelfond, and Nelson Rushton.
\newblock Probabilistic reasoning with answer sets.
\newblock {\em Theory and Practice of Logic Programming}, 9(1):57--144, 2009.

\bibitem{brafman2002r}
Ronen~I Brafman and Moshe Tennenholtz.
\newblock R-max-a general polynomial time algorithm for near-optimal
  reinforcement learning.
\newblock {\em Journal of Machine Learning Research}, 3(Oct):213--231, 2002.

\bibitem{chermack2004improving}
Thomas~J Chermack.
\newblock Improving decision-making with scenario planning.
\newblock {\em Futures}, 36(3):295--309, 2004.

\bibitem{chitnis2018integrating}
Rohan Chitnis, Leslie~Pack Kaelbling, and Tom{\'a}s Lozano-P{\'e}rez.
\newblock Integrating human-provided information into belief state
  representation using dynamic factorization.
\newblock {\em arXiv preprint arXiv:1803.00119}, 2018.

\bibitem{dvzeroski2001relational}
Sa{\v{s}}o D{\v{z}}eroski, Luc De~Raedt, and Kurt Driessens.
\newblock Relational reinforcement learning.
\newblock {\em Machine learning}, 43(1-2):7--52, 2001.

\bibitem{ferreira2017answer}
Leonardo~A Ferreira, Reinaldo~AC Bianchi, Paulo~E Santos, and Ramon~Lopez
  de~Mantaras.
\newblock Answer set programming for non-stationary markov decision processes.
\newblock {\em Applied Intelligence}, 47(4):993--1007, 2017.

\bibitem{hanheide2017robot}
Marc Hanheide, Moritz G{\"o}belbecker, Graham~S Horn, Andrzej Pronobis,
  Kristoffer Sj{\"o}{\"o}, Alper Aydemir, Patric Jensfelt, Charles Gretton,
  Richard Dearden, Miroslav Janicek, et~al.
\newblock Robot task planning and explanation in open and uncertain worlds.
\newblock {\em Artificial Intelligence}, 247:119--150, 2017.

\bibitem{jiang2018empirical}
Yuqian Jiang, Shiqi Zhang, Piyush Khandelwal, and Peter Stone.
\newblock An empirical comparison of pddl-based and asp-based task planners.
\newblock {\em arXiv preprint arXiv:1804.08229}, 2018.

\bibitem{kaelbling1998planning}
Leslie~Pack Kaelbling, Michael~L Littman, and Anthony~R Cassandra.
\newblock Planning and acting in partially observable stochastic domains.
\newblock {\em Artificial Intelligence}, 101(1):99--134, 1998.

\bibitem{khandelwal2017bwibots}
Piyush Khandelwal, Shiqi Zhang, Jivko Sinapov, Matteo Leonetti, Jesse Thomason,
  Fangkai Yang, Ilaria Gori, Maxwell Svetlik, Priyanka Khante, Vladimir
  Lifschitz, J.~K. Aggarwal, Raymond Mooney, and Peter Stone.
\newblock Bwibots: A platform for bridging the gap between ai and human–robot
  interaction research.
\newblock {\em The International Journal of Robotics Research},
  36(5-7):635--659, 2017.

\bibitem{leonetti2016synthesis}
Matteo Leonetti, Luca Iocchi, and Peter Stone.
\newblock A synthesis of automated planning and reinforcement learning for
  efficient, robust decision-making.
\newblock {\em Artificial Intelligence}, 241:103--130, 2016.

\bibitem{mcdermott1998pddl}
Drew McDermott, Malik Ghallab, Adele Howe, Craig Knoblock, Ashwin Ram, Manuela
  Veloso, Daniel Weld, and David Wilkins.
\newblock Pddl-the planning domain definition language.
\newblock 1998.

\bibitem{narvekar2017autonomous}
Sanmit Narvekar, Jivko Sinapov, and Peter Stone.
\newblock Autonomous task sequencing for customized curriculum design in
  reinforcement learning.
\newblock In {\em Proceedings of the 26th International Joint Conference on
  Artificial Intelligence (IJCAI)}, volume 147, page 149, 2017.

\bibitem{oh2015toward}
Jean~H Oh, Arne Supp{\'e}, Felix Duvallet, Abdeslam Boularias, Luis~E
  Navarro-Serment, Martial Hebert, Anthony Stentz, Jerry Vinokurov, Oscar~J
  Romero, Christian Lebiere, et~al.
\newblock Toward mobile robots reasoning like humans.
\newblock In {\em AAAI}, pages 1371--1379, 2015.

\bibitem{puterman1994markov}
Martin~L Puterman.
\newblock {\em Markov decision processes: discrete stochastic dynamic
  programming}.
\newblock John Wiley \& Sons, 1994.

\bibitem{richardson2006markov}
Matthew Richardson and Pedro Domingos.
\newblock Markov logic networks.
\newblock {\em Machine learning}, 62(1):107--136, 2006.

\bibitem{sridharan2015refinement}
Mohan Sridharan, Michael Gelfond, Shiqi Zhang, and Jeremy Wyatt.
\newblock A refinement-based architecture for knowledge representation and
  reasoning in robotics.
\newblock {\em arXiv preprint arXiv:1508.03891}, 2015.

\bibitem{sridharan2017can}
Mohan Sridharan, Ben Meadows, and Rocio Gomez.
\newblock What can i not do? towards an architecture for reasoning about and
  learning affordances.
\newblock In {\em International Conference on Automated Planning and Scheduling
  (ICAPS), Pittsburgh, USA}, pages 18--23, 2017.

\bibitem{sutton1998reinforcement}
Richard~S Sutton and Andrew~G Barto.
\newblock {\em Reinforcement learning: An introduction}.
\newblock MIT press Cambridge, 1998.

\bibitem{tenorth2013knowrob}
Moritz Tenorth and Michael Beetz.
\newblock Knowrob: A knowledge processing infrastructure for cognition-enabled
  robots.
\newblock {\em The International Journal of Robotics Research}, 32(5):566--590,
  2013.

\bibitem{yang2018peorl}
Fangkai Yang, Daoming Lyu, Bo~Liu, and Steven Gustafson.
\newblock Peorl: Integrating symbolic planning and hierarchical reinforcement
  learning for robust decision-making.
\newblock {\em arXiv preprint arXiv:1804.07779}, 2018.

\bibitem{Young2013POMDP}
Steve Young, Milica Gašić, Blaise Thomson, and Jason~D. Williams.
\newblock Pomdp-based statistical spoken dialog systems: A review.
\newblock {\em Proceedings of the IEEE}, 101(5):1160--1179, 2013.

\bibitem{zambaldi2018relational}
Vinicius Zambaldi, David Raposo, Adam Santoro, Victor Bapst, Yujia Li, Igor
  Babuschkin, Karl Tuyls, David Reichert, Timothy Lillicrap, Edward Lockhart,
  et~al.
\newblock Relational deep reinforcement learning.
\newblock {\em arXiv preprint arXiv:1806.01830}, 2018.

\bibitem{zhang2017dynamically}
Shiqi Zhang, Piyush Khandelwal, and Peter Stone.
\newblock Dynamically constructed {(PO)MDPs} for adaptive robot planning.
\newblock In {\em Proceedings of the Thirty-First {AAAI} Conference on
  Artificial Intelligence}, pages 3855--3863, 2017.

\bibitem{zhang2015corpp}
Shiqi Zhang and Peter Stone.
\newblock {CORPP}: Commonsense reasoning and probabilistic planning, as applied
  to dialog with a mobile robot.
\newblock In {\em Twenty-Ninth AAAI Conference on Artificial Intelligence},
  2015.

\end{thebibliography}
}

\end{document}